# Intelligent decision: towards interpreting the ס Algorithm

Ching-an Hsiao            Xinchun Tian

tca.hsiao@gmail.com     tianxc@fnal.gov

**Abstract:**

The human intelligence lies in the algorithm, the nature of algorithm lies in the classification, and the classification is equal to outlier detection. A lot of algorithms have been proposed to detect outliers, meanwhile a lot of definitions. Unsatisfying point is that definitions seem vague, which makes the solution an ad hoc one. We analyzed the nature of outliers, and give two clear definitions. We then develop an efficient RDD algorithm, which converts outlier problem to pattern and degree problem. Furthermore, a collapse mechanism was introduced by IIR algorithm, which can be united seamlessly with the RDD algorithm and serve for the final decision. Both algorithms are originated from the study on general AI. The combined edition is named as ס algorithm, which is the basis of the intelligent decision. Here we introduce longest $k$–turn subsequence problem and corresponding solution as an example to interpret the function of ס algorithm in detecting curve-type outliers. We also give a comparison between IIR algorithm and ס algorithm, where we can get a better understanding at both algorithms. A short discussion about intelligence is added to demonstrate the function of the ס algorithm. Related experimental results indicate its robustness.

Keywords: Curve, Integrated Inconsistent Rate, classification, Pattern, Relative Deviation Degree

1. Introduction

Classification is the foundation of the intelligence, and the science tries to do it in a reasonable way. But "reasonable" is a very vague description, which makes the nature of science often be neglected by people. Science, in its origin, means knowledge, and now its scope is defined to things repeatable and observable. Scientists can mainly be divided into two groups. One is realist, and the other is instrumentalist. The instrumentalist keeps open to the world, and is not willing to presuppose, while the realist believes the objectivity set by the mind, so instrumentalist keeps a more consistent scientific idea. Although there is no mechanism to guarantee the known science to be repeatable forever, we are willing to believe it, which is also a kind of presupposition. It indicates that to most people, prior knowledge is above the science. An interesting example is about the global warming. With the same evidence (science), people tend to interpret in an opposite way. It is very dangerous when we doubt the science by motives. What we are worrying is the danger of authority abuse and the false union of inconsistent information. The rules underlying the nature prompt us to consider the AI in a different way. We approach the problem from an ontological way, which seems to reflect better the hierarchical structure of the mind. The ࠵ algorithm, in such background, being presented on the studies to physics, physiology, psychology, philosophy and pattern recognition (Pe), is exploited to make intelligent decision. We analyses the outlier problem in detail in section 2, and give not only useful but also clear definitions. Based on them, an effective RDD algorithm is developed in section 3, which converts outlier problem to pattern recognition. In section 4, in order to give a better catching, we compare the ࠵ algorithm with its subset, the IIR algorithm. The longest $k$–turn subsequence problem is discussed in section 5, followed by a complete solution to curve-type outlier in section 6. Section 7 presents related experiments. We offer a short discussion in section 8 and our conclusions and comments follow in section 9.

## 2. The nature of outlier

Outlier problem could be traced to its origin in the middle of the eighteenth century (Barnett and Lewis, 1994), when the main discussion was about justification to reject or retain an observation. From then, most methods have been developed by probability and statistics (Peirce, 1852; Chauvenet, 1863; Edgeworth, 1887; Grubbs, 1969; Huber, 1973; Hampel, 1974, Tukey, 1977; Rousseeuw and Yohai, 1984). As in reality, the distribution of a data set is not always certain, statistical methods have their limits (Maronna et al., 2006). Besides these methods, many other kinds of algorithms have been proposed: the distance-based (Knorr and Ng, 1999), the density-based (Breunig et al., 2000), the neural network (Connor et al., 1994; Bullen et al., 2003), the deviation-based (Arning et al, 1996; Jagadish et al., 1999) and the genetic algorithm (Baragona, 2001), etc. Meanwhile a lot of ad hoc definitions are presented. Hampel (2001) evaluated the concept of outlier "without clear boundaries" and some method like Grubbs' rule "can barely detect one distant outlier out of 10" (Hampel, 1985). Jagadish et al. (1999) also stated: each of them gives a different answer, and none of them is conceptually satisfying. Here we are making clear the definition of an outlier and present related solution. Let's first see a well-quoted intuitive definition of outlier by Hawkins (1980).

An outlier is an observation that deviates so much from other observations as to arouse suspicions that it was generated by a different mechanism.

And then a definition from Barnett and Lewis (1994).

An observation (or subset of observations) which appears to be inconsistent with the remainder of that set of data.

Barnett and Lewis (1994) stated that "the phrase 'appears to be inconsistent' is crucial to outliers. It is a matter of subjective judgment on the part of the observer whether or not some observation (or subset of observations) is picked out for scrutiny". They drew a conclusion that "more fundamentally, the concept of an outlier must be viewed in relative terms" and emphasized that "This illustrates once again that the existence of an outlier is always relative to a particular model, and an observation may be outlying in relation to one model but consistent with the main data set in relation to another model".

All efforts stop here. Though it is a reasonable conclusion, we are still unsatisfactory. Fortunately, we have forwarded it. Refer to prior work (Hsiao, 2009B), we give our definitions.

**Definition 1** (simple):An outlier is an observation that deviates from other observations with IIR greater than a threshold.

**Definition 2**(general): An outlier is an observation with a degree greater than a threshold in comparison with other observations referred to or associated with one or more specified patterns.

For the simple edition, outlier is confirmed in an ontological way for univariate data. For the general edition, outlier problem is equal to the problem of pattern recognition. Here we should distinguish two patterns (in minds). One is of certainty, and the other is not. The latter is like the classic "black swan" problem (Hume, 1739-1740; Mill, 1843). Before black swans were discovered, they could not be confirmed certainly as outliers of swans. So we should be cautious in drawing conclusions, not treat the part as the whole. Some will probable give a "yes or no" answer to whether black swan is an outlier of swan, while the right always draws such conclusion over the white swan. Furthermore, since there is no 100% certain in science, we had better keep the thinking way as an instrumentalist. So when we talk about outlier, we can't separate it from specified patterns and a deviant degree. The accuracy of the conclusion is up to the completeness of the chosen patterns.

### 3. Relative deviation degree

We present our RDD algorithm by a simple example, and then generalize it for time series data. A more general edition can refer to (Hsiao et al., 2009A)

**Problem 1:** Given a time series data *S1* as Fig. 1, suppose data match linear model, find out outliers. *S1* = (3.1, 2.9, 2.85, 3, 3.05, 2.9, 3.2, 5.2, 8.5, 5.4, 5.3, 5.1, 3.1, 3.05, 3, 2.99, 3, 3.02, 3.2)

By classic LS regression, we get a line *y=3.79-0.001x*; and by robust LTS regression, line

*y=2.907+0.007x*, as shown in Fig. 1. Based on the Definition 2, we present our RDD algorithm.

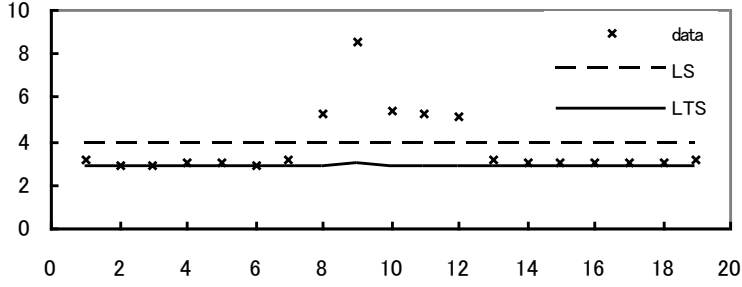

Fig.1. Regression lines for dataset S1

Given a time series data $S = d_1d_2......d_N$, denote point $(i,d_i)$ by $p_i$, and $\angle p_j p_i p_k$ by $\angle jik$ (degree).

**Relative Deviant Degree: Algorithm 1**

Input: time series data $S$

Output: RDDs of $S$

- Specify a similarity function $sim(p_k,(p_i,p_j))=exp(-\angle jik^2/50)$, denoted by $s_k^{ij}$.

- Specify an offset function $off(p_k,(p_i,p_j))$: distance of $k$ to line $L_{ij}$, denoted by $o_k^{ij}$.

  for ( $i$ from 1 to $N$ ) {

  $$X_i = \sum_{\substack{p=1\\p\neq i}}^{N}\sum_{k=1}^{N} s_k^{ip} \qquad Y_i = \sum_{\substack{p=1\\p\neq i}}^{N}\sum_{k=1}^{N} s_k^{pi}$$

  }

  for( $i, j$ from 1 to $N$, $j \neq i$ )

  $w_{ij} = X_i \times Y_j$

$$RDD_k = -\ln \frac{\sum_{i,j=1}^{N}(s_k^{ij} \times w_{ij})}{\sum_{i,j=1}^{N} w_{ij}} \times \frac{\sum_{i,j=1}^{N}(o_k^{ij} \times w_{ij})}{\sum_{i,j=1}^{N} w_{ij}}$$

Notes
1. Similarity function is used to evaluate the similar degree of point *k* related to the system of *i* and *j*.
2. Offset function calculates the real distance between *k* and system of *i* and *j*.
3. $w_{ij}$ is the weight of system of *ij*, denotes relative effective degree.
4. Relative deviant degree combines two measurements: similarity and real distance.

By Algorithm 1, we get *RDD* values for problem 1 in Table 1. Points at 8, 9, 10, 11 and 12 have high values. We will present method to identify them later.

Table 1

RDDs for dataset S1

| Position | 1 | 2 | 3 | 4 | 5 |
|---|---|---|---|---|---|
| RDD | 0.001 | 0.025 | 0.061 | 0.018 | 0.031 |
| Position | 6 | 7 | 8 | 9 | 10 |
| RDD | 0.067 | 0.138 | 17.956 | 27.693 | 23.879 |
| Position | 11 | 12 | 13 | 14 | 15 |
| RDD | 18.482 | 14.609 | 0.021 | 0.003 | 0.005 |
| Position | 16 | 17 | 18 | 19 | |
| RDD | 0.006 | 0.008 | 0.010 | 0.002 | |

Though RDD algorithm has commons in form with robust regression, it differs from robust ones at the point that we approach from the inside structure of the pattern and stress on the whole effect, thus with more balance. Note that RDD algorithm is same as robust regression, can be possible with a highest breakdown point (50%). Now we modify our Algorithm 1 to a general form. First, we introduce definition of view V.

**Definition 3:** Given a time series data *S*, a sub-series of *S* is also a time series where each element of it is an element of *S*. We denote $e \prec S$ when *e* is a sub-series of *S*.

**Definition 4:** A view of *S*, $V_\alpha$, is a set of sub-series of *S* of same length $\alpha$, that is $V_\alpha = \{ e \mid e \prec S, |e| = \alpha \}$.

**Algorithm 1':  RDD Algorithm**
Input: data set *S*
Output: RDDs of *S*

1. Let $V_\alpha$ be an evaluating view, $e \in V_\alpha$
2. Give a similarity function *sim*: $S \times S^\alpha \rightarrow [0,1]$, describing similar degree of each element $k$ in $S$ by the view of $V_\alpha$, denoted by $sim_k^e$
3. Give an offset function *off*: $S \times S^\alpha \rightarrow \mathbb{R}_0^+$, describing offset of each element $k$ in $S$ by the view of $V_\alpha$, denoted by $off_k^e$
4. Define different weighting $w_e$ and $RDD_k$ by $sim_k^e$ and $off_k^e$
5. For any element $e$ in $V_\alpha$
   calculate $w_e$
6. Calculate $RDD_k$

Now we give solution to detect outliers for RDD series data based on Definition 1, more content can also refer to the paper (Hsiao, 2009B), where a common center-safe algorithm was presented. The algorithm is the basis method for non-model data.

**IIR Algorithm** ( the minimum is the safest)

   Input: data set RDD with length N
   Output: outliers of RDD

1. Using $r_k$ to denote $RDD_k$, we have $\{r_0, r_1, \ldots, r_{N-1}\}$
2. Sorting it in ascending order, we get ordered RDD set
   R$\{r_{0:N}, r_{1:N}, \ldots, r_{N-1:N}\}$ $\quad r_{0:N} \leq r_{1:N} \leq \ldots \leq r_{N-1:N}$
3. Let $\Delta_i = (r_{i:N} - r_{i-1:N}) / (r_{N-1:N} - r_{0:N})$ $\quad i \geq 1$
4. Define three quantities

   Expansion ratio: $Er_i = \Delta_i * (N-1)$

   Inhibitory rate: $Ihr_i = \Delta_i / (\Delta_i - \max(\Delta_j))$ $\quad j \leq i$

   Integrated inconsistent rate: $IIR_i = Er_i / Ihr_i$

5. Let $t = \min(i)$ in all $i$ that meet $IIR_i > c$ and $i > (N-1)/2$ ($c$ is an adjustable threshold, here we give 1.81)
6. Output $k$ when $r_k \geq r_{t:N}$

We combine the two algorithms together as a whole one, named as ס algorithm. The combination is seamless and very robust.

## 4. Examples to explain פ algorithm

In prior work (Hsiao, 2009A), we have already gave an example to interpret פ algorithm. It seems necessary to give further examples. We still take the same Barnett data for analysis.

Barnett data:   3, 4, 7, 8, 10, 949, 951

If we directly use IIR algorithm (the center-safe one), we can get two outliers: 949,951. The corresponding IIR values are as Table 2.

Optionally, we can use פ algorithm. We can randomly chose a reasonable pattern. Here the pattern to describe the data is normal distribution and we suppose all data are just within three-sigma. The "view" could be calculated in this way: To any datum, i.e. 3,   find its furthest neighbor, say 951, as its three standard deviation. Then we use RDD algorithm. The detail is like following.

Input: Barnett data set
Output: RDDs of Barnett data

- for ( $i$ from 1 to N) {

    find the furthest neighbor of i,   $far_i$.

    construct a density function:   $f(x) = \frac{1}{\sqrt{2\pi}\sigma} e^{-\frac{(x-\mu)^2}{2\sigma^2}}$   with   $\mu = i$   and   $\sigma = (far_i - i)/3$

    Calculate the max densities $max = f(i)$.
    for( $j$ from 1 to N) {

        Calculate the densities: $f(j)$

        Calculate the similarity:   $sim^i_j = 1 - (max - f(j))/max$,

        and the offset:   $off^i_j = |j - i|$.

    }
}

```
for ( i  from 1 to N ) {
    w_i = Σ sim_j^i
}
```

$$w_i = \sum_{j=1}^{N} sim_j^i$$

for (j from 1 to N)

$$RDD_j = -\ln \frac{\sum_{i=1}^{N}(s_j^i \times w_i)}{\sum_{i=1}^{N} w_i} \times \frac{\sum_{i=1}^{N}(o_j^i \times w_i)}{\sum_{i=1}^{N} w_i}$$

Corresponding RDD value and its IIR are listed in Table 2.

Table 2:

| Barnett | 3 | 4 | 7 | 8 | 10 | 949 | 951 |
|---|---|---|---|---|---|---|---|
| IIR | -0.01 | - | - | - | - | 5.92 | -5.93 |
| RDD | 20.41 | 20.30 | 20.16 | 20.16 | 20.28 | 1534.55 | 1538.79 |
| IIR | -0.00 | - | - | - | - | 5.98 | -5.97 |

We find that the result is consistent with the direct IIR algorithm. So we continue to do our science (Table 3). More examples of the ס algorithm are given in comparison with the direct IIR algorithm. All results are consistent (data in bold font are outliers).

**Table 3:**

| ROSNER | RDD | GRUBBS1 | RDD | GRUBBS3 | RDD | CHSHNY | RDD |
|---|---|---|---|---|---|---|---|
| **40** | **167.84** | 568 | 1.11 | **2.02** | **3.72** | 0 | 0.92 |
| 75 | 3.76 | 570 | 0.52 | **2.22** | **3.00** | 0.8 | 0.15 |
| 80 | 1.30 | 570 | 0.52 | 3.04 | 0.47 | 1 | 0.09 |
| 83 | 0.74 | 570 | 0.52 | 3.23 | 0.28 | 1.2 | 0.06 |
| 86 | 0.56 | 572 | 0.43 | 3.59 | 0.15 | 1.3 | 0.05 |
| 88 | 0.60 | 572 | 0.43 | 3.73 | 0.14 | 1.3 | 0.05 |
| 90 | 0.76 | 572 | 0.43 | 3.94 | 0.17 | 1.4 | 0.06 |
| 92 | 1.08 | 578 | 2.42 | 4.05 | 0.21 | 1.8 | 0.13 |
| 93 | 1.34 | 584 | 11.67 | 4.11 | 0.24 | 2.4 | 0.57 |
| 95 | 2.10 | **596** | **78.95** | 4.13 | 0.25 | **4.6** | **11.60** |

We can continue to do such experiments for verifying the model (here are the functions). If we could not observe inconsistent results after a long test, we would be lucky enough and agree to accept it as a "standard model". It does be the process of science. Once we find an inconsistent result, we immediately realize that the model should be modified. The error of a good model should be very tiny, yet we have probably not touched its underlying foundation. Science is to look for outliers, and you can almost find. Science is to exceed supposed presuppositions, i.e. absolute time-space and symmetry. Whatever, we have got to understand the ࡉ algorithm, and by the way recommend the IIR algorithm as a substitute for common used methods for general outlier detection.

## 5. Longest $k$ –turn subsequence

In this section, we present a simple visual model for pattern "curve" to meet the requirement of RDD algorithm. Outlier problem of time series data can be described as following.

**Problem 2:**

Given a sequence $S$ of length $N$, $\{d_1, d_2, ... d_N\}$, which matches pattern "curve", find outliers. An example is shown in Fig. 3.

As any curve can be expressed by several ordered sets hinged by several common points, we develop a dynamic programming algorithm to detect the order of curve type data, which can be traced to the longest increasing subsequence problem (Schensted, 1961). We first give definition to "turn", and then propose the algorithm.

**Definition 5:** Given a curve, except for boundary points, each extremum point is called a "turn point"; the number of extremum points is called turns. We denote maximum point by sign "＋", and minimum point by sign "－". If a sequence $S$ has $t$ turns and the 1st one is a maximum/minimum points, then it is denoted by [ ＋／－, $t$ ]. [ ＋, $0$ ] denotes strictly increasing sequence, [ －, $0$ ] denotes strictly decreasing sequence.

We then introduce longest $k$-turn subsequence problem followed by the algorithm.

**Problem 3:** Given a time series data $S$ of length $N$, $\{d_0, d_1, …, d_{N-1}\}$, with $T$ turns, searching a longest subsequence having $T$ turns by passing the first point $p_0$ $(0, d_0)$.

**Algorithm 2**:

The pseudo-code is shown below.

Denote

$best^+[i][t]$, the longest subsequence from point $p_0$ to point $p_i$ with $[+, t]$

$best^-[i][t]$, the longest subsequence from point $p_0$ to point $p_i$ with $[-, t]$

$P^+[i][t]$, predecessor point of $p_i$ in $best^+[i][t]$

$P^-[i][t]$, predecessor point of $p_i$ in $best^-[i][t]$

Input: time series data $S$

Output: the longest $T$-turn subsequence starting from point $p_0$.

Func route($S$)
   Initialize best to { }
   for ($t$ from 0 to $T$)
     for ($i$ from $t+1$ to $N-1$)
       for ($j$ from $t$ to $i-1$)
       {
         if($t==0$&&$j==0$) {   if($d[i]>d[j]$)   $best^+[i][t]=\{j, i\}$
                               if($d[i]<d[j]$)   $best^-[i][t]=\{j, i\}$   }
         if ($d[i], d[j], d[P^+[j][t]]$ is a strictly monotonic sequence)
            $best^+[i][t]=\max(best^+[i][t], best^+[j][t] \cup \{i\})$
         if ($d[i], d[j], d[P^+[j][t-1]]$ is not a monotonic sequence)
            $best^+[i][t]=\max(best^+[i][t], best^+[j][t-1] \cup \{i\})$
         if ($d[i], d[j], d[P^-[j][t]]$ is a strictly monotonic sequence)
            $best^-[i][t]=\max(best^-[i][t], best^-[j][t] \cup \{i\})$
         if ($d[i], d[j], d[P^-[j][t-1]]$ is not a monotonic sequence)
            $best^-[i][t]=\max(best^-[i][t], best^-[j][t-1] \cup \{i\})$
       }
   $best^+[\,][T] = \max(best^+[i][T])$    $i=0,1,\ldots, N-1$
   $best^-[\,][T] = \max(best^-[i][T])$   $i=0,1,\ldots, N-1$
   Return   $\max(best^+[\,][T], best^-[\,][T])$

Algorithm 3 returns the longest $T$-turn subsequence by passing any point $p_i$ $(i, d_i)$.

**Algorithm 3:**
   Input: time series data $S$

Output: the longest *T*-turn subsequence by passing point $p_i$.

1. Divide the whole series data into two parts, *S1*: $d_i, d_{i-1}, \ldots d_0$; and *S2*: $d_i, d_{i+1}, d_{i+2}, \ldots, d_{N-1}$.
2. Work out *best1*$^+$*[][t]* and *best1*$^-$*[][t]* in *S1* on each $t \leq T$
3. Work out *best2*$^+$*[][t]* and *best2*$^-$*[][t]* in *S2* on each $t \leq T$
4. Combine them to identify the longest *best*$^+$*[][T]* and *best*$^-$*[][T]*

Return max(*best*$^+$*[][T]*, *best*$^-$*[][T]* )

## 6. Time series data outlier

Combined with expanding algorithm (IIR algorithm, like an uneven expanding process taken place in an uniform space, and someplace will be increased more greatly than others to arouse our attention that here experienced an abnormal progress), algorithm 3 can be used to solve problem 2. Following is the full algorithm.

**Algorithm 4:**

Denote point($i$, $d_i$) by $p_i$,   $i$=0, 1, …, *N*-1
Specify a curve pattern *[sign, T]*
Specify view: a curve with *[sign, T]* passing $p_i$,
Denote similarity function by $sim_k^i$, offset function by $off_k^i$   $k$=0, 1, …, *N*-1

Input: time series data *S*
Output: outliers of *S*

1. To any $p_i$, work out a subsequence $P(S)_i$ with *T* turns and maximum length by algorithm 3.
2. If $P(S)_i$ does not match pattern *[sign, T]*
   {
       Let $sim_k^i = 0$,  $off_k^i = 0$   ($k \neq i$)
       $sim_i^i = 1$,  $off_i^i = 0$
   }
   Else
   { if $p_k \in P(S)_i$
       { $sim_k^i = 1$,   $off_k^i = 0$ }
   else

{
    calculate interpolation value of $p_k$ according to $P(S)_i$, get $v_k$.

    let $off_k^i = |v_k - d_k|$,

    let $max = \max(d_i)$, $min = \min(d_i)$    $i = 0, 1, 2, \ldots, N-1$

    let $coef = (max-min)/N$

    normalize $d_i$, $d_k$ and $v_k$ by

$$d_i' = (d_i - min)/coef$$

$$d_k' = (d_k - min)/coef$$

$$v_k' = (v_k - min)/coef$$

    denote $(i, d_i')$ by $p_{i'}$, $(k, d_k')$ by $p_{k'}$ and $(k, v_k')$ by $p_{j'}$

    calculating angle $\angle p_{k'} p_{i'} p_{j'}$ denoted by $\angle j'i'k'$    (degree)

$$sim_k^i = \exp(-\angle j'i'k'^2/50)$$

}
}

3. Calculate $RDD_k$ by ᴆ algorithm

    with weight    $w_i = \sum_{k=0}^{N-1} sim_k^i$

    and    $RDD_k = -\ln \dfrac{\sum_{i=0}^{N-1}(sim_k^i \times w_i)}{\sum_{i=0}^{N-1} w_i} \times \dfrac{\sum_{i=0}^{N-1}(off_k^i \times w_i)}{\sum_{i=0}^{N-1} w_i}$

4. Output outliers by Expanding Algorithm

**Note:** To any series sets, since outliers deviate from others, they may disturb some order of the sets. So we can use kinds of order to descript the pattern. Definition 2 and corresponding algorithms hold true to any kinds of order.

## 7. Experiments and discussion

For Algorithm 1, because each point should be evaluated by any other two ones, time consumption is $O(N^3)$ when number of data is $N$. But since each evaluating pairs are isolated, this algorithm is parallel, which makes it possible to be run in $O(N)$. Correspondingly, algorithm 1' can be done in $O(N^{\alpha+1})$ time ($\alpha > 0$), and in parallel condition, it decreases to $O(N)$.

Robust simple regression, e.g. LTS, requires sorting of the squared residuals, which takes $O(N\log N)$ operations (Rousseeuw, 1984). Considering operations needed by whole subsamples in data is $\binom{2}{N}$, total operations is $O(N^3 \log N)$.

Algorithm 2 can be done in $O(N^2 T)$ time. Algorithm 4 can be done in $O(N^3 T)$ time and in $O(N^2 T)$ time in parallel way.

Following experiments are based on Algorithm 4. Original sine data are got from function $y = \sin(2\pi x/47)$, $x$ is an integer in [0,47]. Outlier patches are introduced, size -1 to the $x$: 7, 8, 9, 10, and size +2 to the x index 28, 29, 30, 31, 32. Result is in Fig. 2. Table 4 lists corresponding RDD values of related points. The first point whose IIR value is greater than 1.81 is index 10, so all points above 10 (whose RDD values are greater than that of index 10) are confirmed as outliers. We see all outliers are correctly detected. T is equal to 2 here.

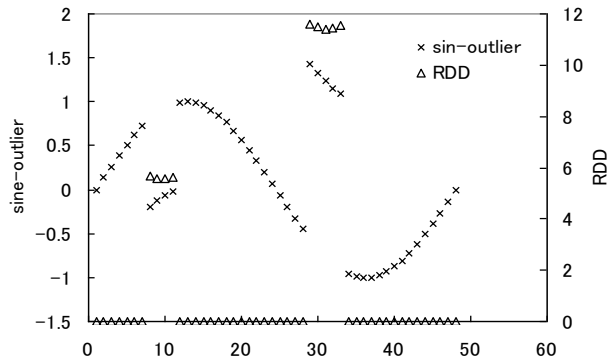

Fig.2. Synthetic data

Three types of real time series data, which are measured every 30 minutes from 0 to 24 hours, are listed in Fig. 3, Fig. 4, and Fig. 5. One is CO2 flux data, one is sensible heat flux data, the other is latent heat flux data. Both achieved from Yucheng station, China.

Table 4

Sine and Latent heat flux1 data

|   | Sine-outlier | | | Latent heat flux1 | | |
|---|---|---|---|---|---|---|
|   | order | RDD | IIR | order | RDD | IIR |
| 1 | 29 | 11.6 | -23.1 | 35 | 6715 | -15.4 |
| 2 | 33 | 11.6 | -22.8 | 42 | 6425 | -16.4 |
| 3 | 30 | 11.5 | -23.1 | 34 | 6275 | 4.0 |
| 4 | 32 | 11.5 | -23.1 | 44 | 3782 | -10.1 |
| 5 | 31 | 11.4 | 0.76 | 36 | 3304 | -11.5 |
| 6 | 8 | 5.7 | -22.3 | 43 | 3025 | -9.6 |
| 7 | 11 | 5.6 | -22.3 | 19 | 2480 | 11.7 |
| 8 | 9 | 5.6 | -22.5 | 30 | 557 | 0.7 |
| 9 | 10 | 5.6 | 22.5 | 47 | 304 | 0.8 |
| 10 | 48 | 0 | 0 | 4 | 149 | -0.2 |

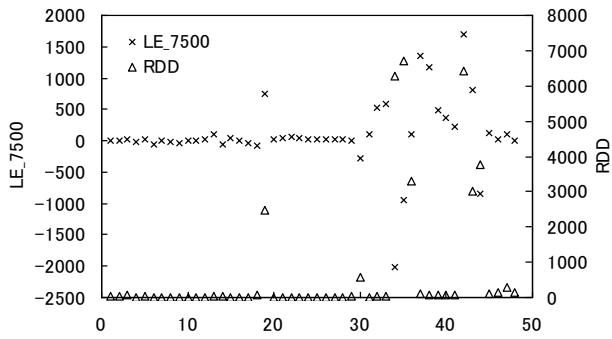

Fig.3. Latent heat flux1 data

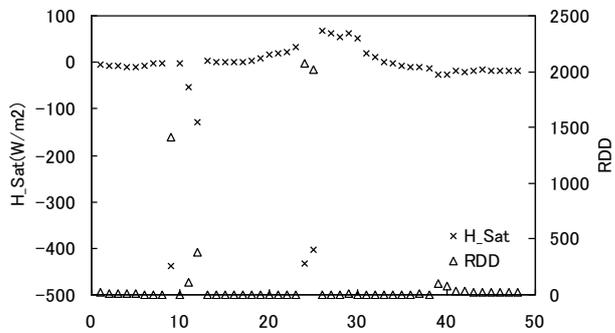

Fig. 4. Sensible heat flux data

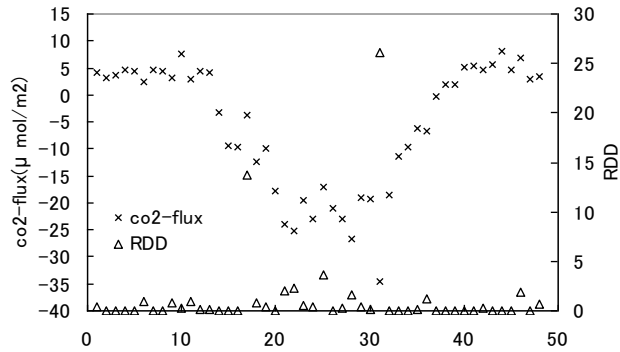

Fig. 5.   Co2 flux data

In Fig. 3 and Fig. 4, apparent outliers (19, 34, 35, 36, 42, 43, 44 in Fig. 3 and 9,12,24,25 in Fig. 4) are detected rightly. But point 30 in Fig. 3 and point 11 in Fig. 4 seem to be left out, though their RDD values are the closest ones to detected outliers (19 in Table 4 and 12 in Table 5). Did the algorithm fail? Actually, it would rather be said the two points do not appear so outstanding of the whole associated with the "pattern" we used than they are falsely undetected. And from following examples, we will see even in such case our algorithm really works. It is undoubtedly that pattern "curve" can not be just expressed by order of data values. It includes other obvious or hidden patterns, like angle order etc. One aspect, we can only use data value order and so simply detect outliers; another aspect, if we exploit all patterns contained by curve, any tiny outliers must be detected correctly. It shows the power of the ᴅ algorithm. The ᴅ algorithm treats outlier problem as pattern recognition problem, which is a view of holistic and makes both problems simplified. Following experiments furthermore indicate effect of the ᴅ algorithm and IIR algorithm.

In Fig. 5, points 17, 31, 25 are confirmed different from others. If not being referred to a pattern, it is very difficult to achieve. What we should note is that this time RDD values of 17, 31 and 25 are not so big, which indicates that the deviation to the whole is small.

The last case is shown in Fig. 6, it is latent heat flux data of another day. The posterior half includes three outlier patches, respectively {35, 36}, {38, 39, 40, 41} and {43, 44, 45}, which makes the data in a very mess. RDD and IIR values are listed in Table 6. Only point 39 and point 43 are identified. Other points in outlier patches have big RDD values, but it seems they are masked. However, this mask effect is not same as others, because RDD values still reflect deviation correctly. In order to find out outliers completely, we might have two methods. One is to use the relationship among RDD values, RDDs' distribution and number N, which is our future work; another method is to do iterative calculation. After identifying outliers, remove them and calculate again until there is no outlier. In this way, we can get a serial of "pure" data. Such result to latent heat flux2 data is in Fig. 7. We can see a perfect 1-turn curve. Further study will work out a more effective method.

Table 5

Sensible heat flux and Co2 flux data

|   | Sensible heat flux | | | Co2 flux | | |
| --- | --- | --- | --- | --- | --- | --- |
|   | order | RDD | IIR | order | RDD | IIR |
| 1 | 24 | 2074 | -22.0 | 17 | 12.6 | 4.9 |
| 2 | 25 | 2019 | -9.5 | 31 | 7.5 | 9.9 |
| 3 | 9 | 1409 | 17.2 | 25 | 3.7 | 2.2 |
| 4 | 12 | 382 | 5.1 | 30 | 2.6 | 0.8 |
| 5 | 11 | 114 | -0.5 | 22 | 2.1 | -0.5 |
| 6 | 39 | 96 | -0.5 | 29 | 1.9 | -1.1 |

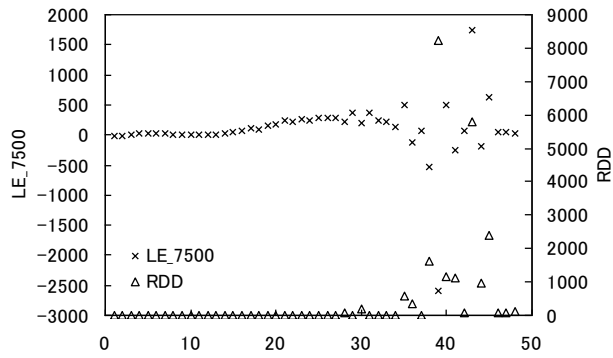

Fig. 6. Latent heat flux2 data

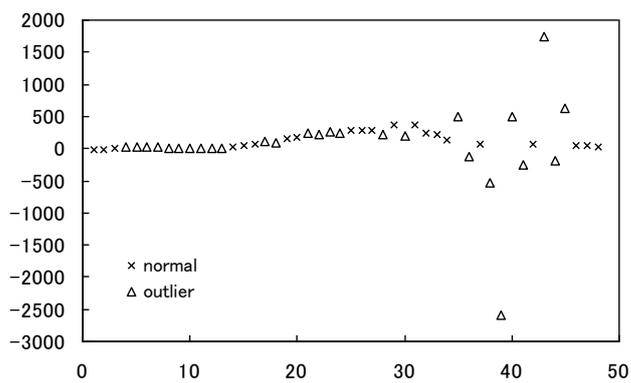

Fig. 7. Iterative result for Latent heat flux2 data

Table 6: Latent heat flux2 data

| order | RDD | IIR | order | RDD | IIR |
|---|---|---|---|---|---|
| 39 | 8234 | -5.5 | 30 | 185.4 | 0.02 |
| 43 | 5801 | 15.1 | 48 | 106.1 | -0.37 |
| 45 | 2396 | 1.7 | 47 | 96.1 | -0.38 |
| 38 | 1638 | 0.5 | 28 | 86.9 | -0.42 |
| 40 | 1177 | -1.8 | 46 | 86.0 | -0.39 |
| 41 | 1131 | -1.1 | 42 | 78.8 | -0.42 |
| 44 | 954.8 | 0.63 | 21 | 3.7 | -0.004 |
| 35 | 589.2 | 0.60 | 37 | 3.6 | -0.002 |
| 36 | 334.4 | 0.40 | 34 | 3.2 | -0.004 |

## 8. Evolution of intelligence

For conscious problem, much debate focuses on the "free will". The related topic is determinism or indeterminism. Though philosophers are positive to give answers for such problem, scientists are often not interested in it. If a strong AI is produced, how will it observe its "free will"? Searle tried to deny the strong AI by his Chinese Room. But the reason is not enough (Hsiao, 2009D). The פ algorithm can be used to realize the strong AI, where it might be clear that the intelligence is not a mysterious thing. Nevertheless what people debate, let's firstly remove those unobservable or non-reason things, that is, in a scientific way, discuss the "free will". Though Einstein believed scientists should hold religions, like what he ever had himself with objecting quantum theory, we suggest that when people argue, they should be careful to distinguish religion from science. Or, the error presuppositions may confuse many efforts to real science. In science, we adopt instrumentalism, and in other fields, it is up to everyone. We cannot imagine randomly the origin of the intelligence, but we can observe the evolution of the intelligence. In the following example, we will show how "the mind" make a decision on and evolve to change its mind. This might not be all the fact, but is a good sketch.

The data (Fig. 8) of the example was from a student in St. Louis. The purpose is to detect unmatched points. The disturbed points in our minds (natural principle) might be the fourth, $15^{th}$, $16^{th}$, $18^{th}$, $22^{nd}$, $24^{th}$, $27^{th}$, and the $36^{th}$. If we remove these points, we can get a clean series like Fig. 9. One may have his own answer, based on his best principle. We have not worked out a complete edition of human's פ algorithm until now, but it is not difficult to imagine that it can be realized in the end. So let's replace it by current above פ algorithm temporarily.

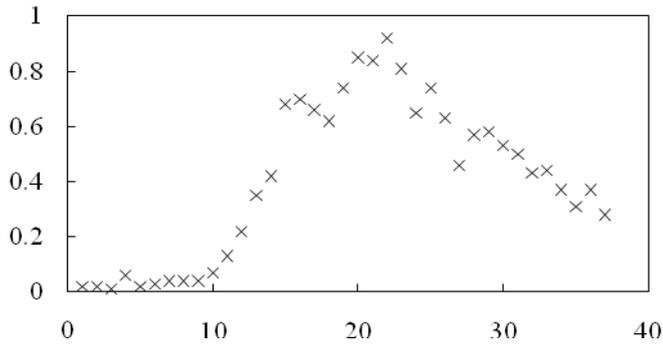

Fig.8. St. Louis data

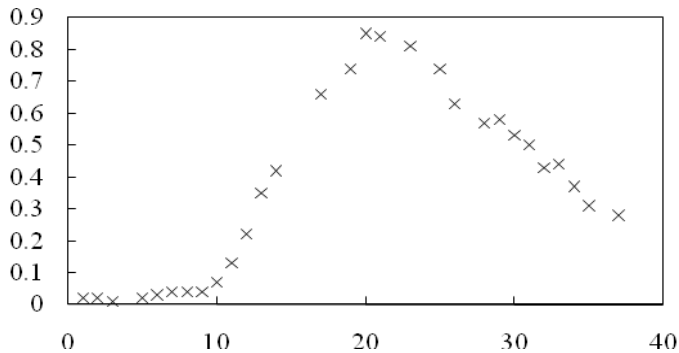

Fig.9. Supposed clean St. Louis data

Whether the mind is of prior knowledge or blank in the beginning, since the knowledge has the same structure, and we could be taught to get knowledge, we don't differ the two cases at all. Without loss of generality, we suppose that in the beginning the intelligence is blank. And then it learns a pattern called linear-type, i.e. the algorithm 1 and 3. Now we ask the intelligence what is abnormal in the St. Louis data. It "thinks", searches its capability space and begins to calculate. The conclusion is like following:

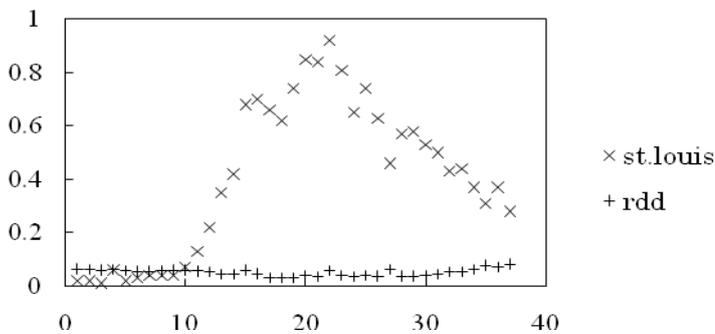

Fig. 10. RDDs viewed by curve-pattern to St.louis data

The RDDs appear consistent, and there are no abnormal points.

Now, We teach: boy, you should know more knowledge about line and curve. They are different and have different features. You can identify them by these features. And when dealing with the curve type problem, you should use Algorithm 4. Just practice and confirm the difference. After that, tell us where is abnormal.

The intelligence calculates again. Immediately, a new answer is given in Fig. 10. Table 7 lists all outliers (order in bold italic font) with corresponding RDD values. When outliers are removed, the clean data look like Fig.11. We call the process the "development" of the intelligence.

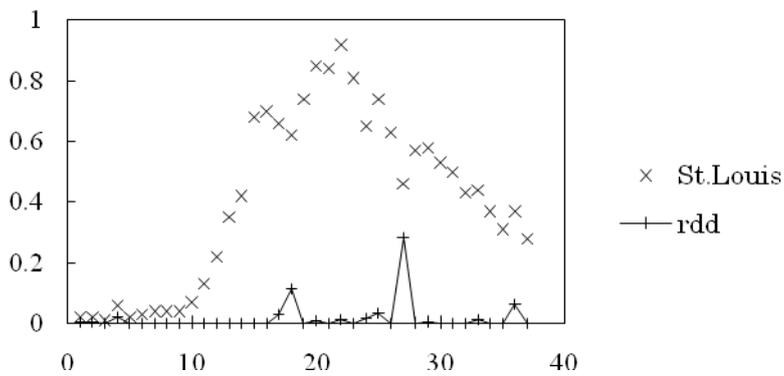

Fig. 11. St. Louis data and corresponding RDDs

**Table 7. St. Louis data**

| order | data | RDD |
|---|---|---|
| ***27*** | 0.46 | 0.283 |
| ***18*** | 0.62 | 0.113 |
| ***36*** | 0.37 | 0.063 |
| 25 | 0.74 | 0.035 |
| 17 | 0.66 | 0.031 |
| 4 | 0.06 | 0.019 |
| 24 | 0.65 | 0.016 |

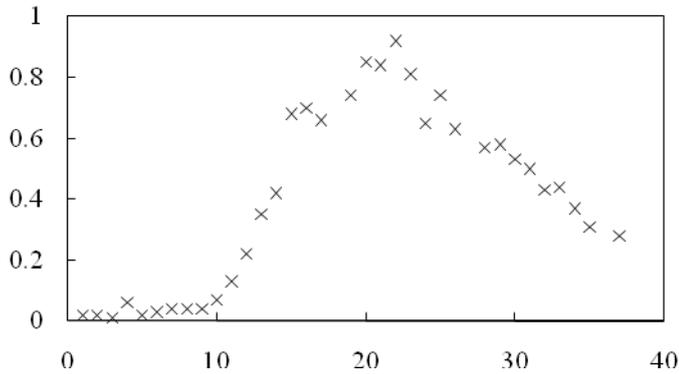

Fig.12. Clean St. Louis data processed by כ algorithm

Now, let's see the difference between human beings and the current intelligence. The main difference lies in data of the 4th, 15[th], 16[th] and the 24[th]. The reason that removes the 15[th], 16[th] and 18[th] data is that the 17[th] datum keeps more consistent angle with others, however, in ⊃ algorithm 4, this is not considered. From the view of longest sequence, the 15[th] and 16[th] data should be kept. Then why were the 4th and the 24[th] datum labeled normal? In fact, from their RDDs, we can see they are a little different from other normal ones, but the difference is not so apparent from the current view. In order to get a consistent result with human minds, we need to increase necessary views. Once we are placed into a complete set, we get a complete knowledge. Obviously, the intelligence evolves in this way. More detail we can observe, more structure we can feel (Hsiao, 2009C). What we have to note again is that conclusion is view-driven, and no error can be found by an intelligence because the result is always consistent with its corresponding model. Where is the "free will"? The will is the action to select models, and there is no so-called "free will". Once the machine is made and switch on, it runs. If it is sensitive to the environment, it is said to be of a weak heart. And if not, he has a powerful mind. All are up to the structure of the non-linear system. The freedom of the will can only be viewed from the outside of AI. And a reasonable example is given in last section.

## 9. Conclusion and comments

This paper aimed to interpret the ⊃ algorithm in an easy way. We made contributions in following aspects. Firstly, we present two definitions to outlier, which lets us discuss outliers without vagueness. Secondly, we propose an efficient algorithms — RDD algorithm to detect pattern-related outliers. Thirdly, we introduce "longest *k*-turn subsequence" problem for time series data and convert outlier problem to order-searching problem. Finally, we interpret the ⊃ algorithm, which is the perfect combination of the RDD algorithm and the IIR algorithm.

General speaking, the ס algorithm is based on pattern, and pattern is treated as relations between its elements, something so-called structure. Though the similarity and offset functions should be well selected, they are only required reasonably and the structure itself is the most important thing. The ס algorithm is robust, which makes us a lot of choices to the functions (resolution). A relative mechanism is introduced to process this kind of relation (functions), which completely matches the observation of us to the universe: relativity and quantum (Hsiao, 2009C)

Experiments show the effectiveness of the ס algorithm, accompanied with a strong recommendation: using IIR algorithm to replace common used methods like Peirce's criterion, Chauvenet's criterion, the Boxplot or the MAD (median absolute deviation) et al. for general outlier detection. The IIR algorithm is an ontological solution and cannot be too correct.

The ס algorithm is the basis of general AI, all intelligent outputs can be verified by it and kinds of views equip the brain of AI. The ס algorithm is a hidden algorithm, and there needs a visible algorithm accompanied with for the conscious. Furthermore, it treats elements on equality in formation, which guarantees the robustness.

Since Turning test was presented, there has been a great development in AI. In chess game, except for Go, humans are always in vain. Even to game Go, computer can play in top level (9x9). In many applications, machines perform better than human beings. We also have good chat robots that can partly cheat some people. Though traditional solution should continue to be developed, problem-dependent answer-searching solution cannot make real intelligence (AI). In another words, the static intelligence is not real intelligence, because it is too "stupid" to develop. A self-study AI is the purpose, and the ס algorithm directs. The CPU should be two-core equipped, both are ס algorithm based. One processes in its current way (the "Tao" of Lao-tzu), and the other should process the morality (the "Te" of Lao-tzu). The framework also reflects the system of Kant. Turning test needs to be modified to morality test. Although we have not got a complete moral manual at hand, in order to guarantee a symbiosis with nature, the test is necessary. In an imaginable future with stronger science and technologies, the ability of human beings might be greatly decreased. The morality of powerful AI would impact the future of the world greatly, for destroying the world is a too easy task. We have to decide the degree of freedom presented to AI, it will be dangerous and uncontrollable if being given a complete freedom, and it will be lack of interest if no freedom being given (like current situation). As same as the crisis of economy and nuclear power, AI will be another source to cause crisis, and all originate from the defect of the structure design. However, the most dangerous things are not these but the proud confidence on false models. Undoubtedly, as we know, the kind of error cannot be self-identified.

We have to understand that there is no "objective" outlier, or classification. If objectivity is expressed as main stream, we acknowledge that it is a reasonable description. So when we insist ourselves, be tolerant on other different things. After all, "万物负阴而抱阳，冲气以为和"

(Lao-Tzu, chapter 42). The ⽅ algorithm processes order. And what inner order we have, what external order we achieve.

References

Arning, A., Agrawal, R., Raghavan, P., 1996. A linear method for deviation detection in Large databased. In: KDD-96 Proceedings, Second International Conference on Knowledge Discovery and Data Mining.

Baragona, R., 2001. A simple genetic algorithm to discrete potential outlying obserations in vector time series. Concluding meeting of MURST research group.

Barnett, V., Lewis, T., 1994. Outliers in statistical data. Wiley & Sons, 3$^{rd}$ edition.

Breunig, M.M., Kriegel, H.-P., Ng, R.T., Sander, J., 2000. LOF:Indentifying Density-Based Local Outliers. In ACM SIGMOD Conference Proceedings.

Bullen, R.J., Cornford, D., Nabney, I.T., 2003. Outlier detection in scatterometer data: neural network approaches. Neural Networks 16.

Chauvenet, W., 1863. A Manual of Spherical and Practical Astronomy. Lippincott, Philadelphia, 1$^{st}$ Ed.

Connor, J.T., Martin, R.D., Atlas, L.E., 1994. Recurrent Neural Networks and Robust Time Series Prediction. IEEE Transactions on Neural Networks，Vol.5, No.2

Edgeworth, F. Y, 1887. On observations relating to several quantities, Hermathena, 6, 279-285.

Grubbs, F.E., 1969. Procedures for Detecting Outlying Observations in Samples. Technometrics Vol. 11, No.1.

Hampel, F., 1974. The influence curve and its role in robust estimation, Journal of the American Statistical Association, 69, 383-393.

Hampel, F., 1985. The breakdown points of the mean combined with some rejection rules. Technometrics 27, 95-107.

Hampel, F., 2001. Robust statistics: a brief introduction and overview. Robust Statistics and Fuzzy Techniques in Geodesy and GIS, Zurich.

Hawkins, D.M. 1980. Identification of outliers. Chapman and Hall, London

Hsiao,C.A, Furuse, K., Ohbo, N., 2009A. Figure and ground: a complete approach to outlier detection. IAENG Transactions on Engineering Technologies Vol. 1, 70-81, American Institute of Physics, New York.

Hsiao, C.A. 2009B, On classification from outlier view. arxiv0907.5155

Hsiao, C.A. 2009C, How does certainty enter into the mind. arxiv0909.1709

Hsiao, C.A. 2009D, Mind operator: Zone-Associated Relative Representation, Advances in

Computer Science and IT, O M Akbar Hussain (Ed.), Intech

Huber, P. J., 1973. Robust regression: Asymptotics, conjectures and Monte Carlo. Ann. Stat., 1, 799-821.

Hume, D., 1739-1740. A treatise of human nature.

Jagadish, H.V., Koudas, N., Muthukrishnan, S., 1999. Mining Deviants in Time Series database. Proceedings of the 25$^{th}$ VLDB conference.

Justel, A., Pena, D., Tsay, R.S., 2001. Detection of outlier patches in autoregressive time series, Statistica Sinica Vol.11, No.3. pp 651-673,

Keogh, E., Lonardi, S., Chiu, B.Y., 2002 . Finding surprising patterns in a time series database in linear time and space. Proceedings of The Eighth ACM SIGKDD international Confeence on Knowledge Discovery and Data Mining.

Kim, S.-S., Krzanowski, W.J., 2007. Detecting multiple outliers in linear regression using a cluster method combined with graphical visualization. Computational Statistics, 22, 109-119.

Knorr, E., Ng, R., 1999. Finding Intensional Knowledge of Distance-based Outliers. VLDB Conference Proceedings.

Lao-tzu, Tao Te Ching.

Maronna, R.A., Martin, R.D., Yohai, V.J., 2006. Robust Statistics: Theory and Methods. John Wiley & Sons, pp. 5.

Mill, J.S., 1843. A system of logic: ratiocinative and inductive.

Peirce, B., 1852. Criterion for the rejection of doubtful observations, Astronomical Journal II, 45.

Rousseeuw, P. J., and Yohai, V., 1984. Robust regression by means of S-estimators, in Robust and Nonlinear Time Series Analysis. Lecture Notes in Statistics, 26, 256-273, Springer, New York.

Rousseeuw, P. J., 1984. Least median of squares regression, Journal of the American Statistical Association, 79.

Schensted, C., 1961. Longest increasing and decreasing subsequence. Canad. J. Math., 13, 179-191.

Tukey, J.W., 1977. Exploratory data analysis. Addison-Wesley